# A New Distributed Evolutionary Computation Technique for Multi-Objective Optimization

Md. Asadul Islam[1], G.M. Mashrur-E-Elahi[2], M.M.A. Hashem[3]
Department of Computer Science and Engineering
Khulna University of Engineering & Technology (KUET)
Khulna 9203, Bangladesh
asad_kuet@yahoo.com[1], ranju2k4cse_kuet@yahoo.com[2], mma_hashem@hotmail.com[3]

*Abstract*— **Now-a-days, it is important to find out solutions of Multi-Objective Optimization Problems (MOPs). Evolutionary Strategy helps to solve such real world problems efficiently and quickly. But sequential Evolutionary Algorithms (EAs) require an enormous computation power to solve such problems and it takes much time to solve large problems. To enhance the performance for solving this type of problems, this paper presents a new Distributed Novel Evolutionary Strategy Algorithm (DNESA) for Multi-Objective Optimization. The proposed DNESA applies the divide-and-conquer approach to decompose population into smaller sub-population and involves multiple solutions in the form of cooperative sub-populations. In DNESA, the server distributes the total computation load to all associate clients and simulation results show that the time for solving large problems is much less than sequential EAs. Also DNESA shows better performance in convergence test when compared with other three well-known EAs.**

I. INTRODUCTION

Multi-Objective optimization optimize a set of conflicting objectives simultaneously. MOP is a very important research topic, not only for the Multi-Objective nature of most real-world decision problems, but also there are still many open questions in this area. Traditionally, there are several methods available in the Operational Research (OR) literature for solving MOPs as mathematical programming models [1]. None of the OR methods treats all the objectives simultaneously which is a basic requirement in most MOPs. In addition, these methods handle MOPs with a set of impractical assumptions such as linearity and convexity. In MOPs, there is no single optimal solution, but contains a set of alternative solutions. These solutions are optimal in the wider sense since there is no other solutions in the search space that is superior to them when all objectives are simultaneously considered. They are known as pareto-optimal solutions or non-dominated solutions [2][3]. In principle, multiobjective optimization is difficult than the single-objective optimization. In single objective optimization, one attempts to obtain the best design or decision, which is usually the global minimum or the global maximum depending on the optimization problem.

Recently, EAs are found to be useful for solving MOPs [4]. EAs have some advantages over traditional OR techniques. This allows us to find several members of the Pareto-optimal set in a single run of the algorithm [5]. Also, there is no requirement for differentiability of the objective functions and the constraints. Moreover, evolutionary algorithms are susceptible to the shape of the Pareto front and can easily deal with discontinuous or concave Pareto front. There is no well-accepted method for MOPs that will produce a good set of solutions for all problems. Also, sequential EAs require an enormous computing power and take much time to solve large problems. This motivates the further development of good approaches to MOPs.

This paper proposes a parallel evolutionary algorithm called Distributed Novel Evolutionary Strategy Algorithm (DNESA) for MOPs, which is developed from a sequential evolutionary algorithm called Novel Evolutionary Strategy (NES) algorithm [10]. The proposed DNESA uses parallel approach to find out the member of pareto-optimal solution, which is more promising than sequential approach. Also diversity of population and fast convergence are achieved by using subpopulation and parallelization. In DNESA, the server divides the problem loads into all associate clients and the simulated results show that it takes less time to solve large problems than sequential EAs. For the convergence test, DNESA is compared with three algorithms namely Pareto Enveloped-based Selection Algorithm (PESA) [6], Non-Dominated Sorting Genetic Algorithm (NSGA-II) [7] and the Strength Pareto Evolutionary Algorithm (SPEA-2) [8]. The compared results showed that DNESA achieved better convergence with respect to other three algorithms.

The organization of the paper is as follows. Section II discusses various existing techniques of Multi-Objective optimization. A brief discussion about the proposed approach is presented in section III. Experimental setup and results is presented in section IV. Section V concludes the paper.

II. BACKGROUND INFORMATION

*A. Problem Definition*

The MOPs [9] (also called multicriteria optimization, multiperformance or vector optimization problem) can be

defined as the problem of finding a vector of decision variables *x*, which optimizes a vector function

$$f_m(x); \quad m = 1, 2, 3, \ldots, M$$

satisfies inequality

$$g_j(x) \geq 0; \quad j = 1, 2, \ldots, J;$$

and equality constraints

$$h_k(x) = 0; \quad k = 1, 2 \ldots K;$$

and whose elements represent the objective functions. These functions form a mathematical description of performance criteria, which are usually in conflict with each other. Hence, the term optimization means finding such a solution, which would give the values of all the objective functions acceptable to the decision maker. These objective functions constitute a multi-dimensional space in addition to the usual decision space. This additional space is called the objective space, Z. For each solution *x* in the decision variable space, there exist, a point in the objective space:

$$f(x) = Z = (z_1, z_2, \ldots, z_M)^T$$

In Multi-Objective optimization, the concept of Pareto dominance and Pareto optimality will form the basis of solution quality. Specifically, the important concepts are defined as follows.

Definition 1: Consider without loss of generality the following multiobjective optimization problem with m decision variables x (parameters) and n objectives y:

Minimizes

$$y = f(x) = (f_1(x_1, x_2, \ldots x_m), \ldots f_n(x_1, x_2, \ldots x_m))$$

$$\text{Where } x = (x_1, x_2, \ldots x_m) \in X$$

$$y = (y_1, y_2, \ldots y_m) \in Y$$

and where *x* is called decision (parameter) vector, X parameter space, *y* objective vector and Y objective space. A decision vector $a \in X$ is said to dominate a decision vector $b \in X$ (also written as) $a \succ b$ if and only if:

$$\forall_i \in 1, 2, \ldots, n; \; f_i(a) \geq f_i(b)$$

$$\wedge \exists_j \in 1, 2, \ldots, n; \; f_j(a) \geq f_j(b)$$

Additionally, we say a covers b ($a \succeq b$) if and only if $a \succ b$ or $f(a) = f(b)$.

*B. Existing Approaches*

There are many Multi-Objective Evolutionary Algorithms (MEAs) for solving MOPs. They are classified as plain aggregating, population-based non-Pareto and Pareto-based approaches [1]. According to this classification, various types of EAs for MOPs are discussed here. The Random Sampling Evolutionary Algorithm (RAND) [4] generates randomly a certain number of individuals per generation. RAND generates individual according to the rate of crossover and mutation.

One of the popular population-based non-Pareto approaches is Vector Evaluated Genetic Algorithm (VEGA) [14]. In this approach, the total population is divided into sub populations based on the number of objective functions to be optimized. Each sub population is used to optimize each objective function independently. The crossover and mutation operators are used to shuffle the population together. According to Schaffer the solutions generated by VEGA were non-dominated in a local sense, because their non-dominance was limited to the current population, and while a locally dominated individual is also globally dominated, the converse is not necessarily true. There are also a non-Pareto approach that uses the weighted sum method for fitness assignment which is known as Hajelas and Lins genetic algorithm (HLGA) [15]. A weight between zero and one is assigned to each objective, with the sum of all weight being exactly equal to one. To search for multiple solutions in parallel, the weights are encoded in the genotype. Phenotypic fitness sharing is used to maintain diversity of the weight combinations. The EA evolves solutions and weight combinations simultaneously.

The dominated and non-dominated solutions in the current population are separated in the Pareto-based approaches. Goldberg [16] suggested a non-dominated ranking procedure to decide the fitness of the individuals. After that based on the idea Golgerg, Srinivas and Dev [12] introduced Non-dominated Sorting Genetic Algorithms (NSGA).

In this method, the fitness assignment is carried out through several steps. In each step, the non-dominated solutions constituting a non-dominated frontier are assigned the same dummy fitness value. These solutions have the same fitness values and are ignored in the further classification process. Finally, the dummy fitness is set to a value less than the smallest shared fitness value in the current non-dominated frontier. Then the next frontier is extracted. This procedure is repeated until all individuals in the population are classified. Fonseca and Fleming [17] proposed a slightly different scheme which is known as Fonseca and Fleming's genetic algorithm (FFGA). In this approach, an individual's rank is determined by the number of individuals dominating it.

Horn and Nafpliotis [18] used phenotypic sharing on the objective vectors. The common features of the Pareto-based approaches mentioned above are that (i) the Pareto-optimal solutions in each generation are assigned either the same fitness or rank, and (ii) some sharing and niche techniques are applied in the selection procedure. Recently, Zitler and Thiele [4] proposed a Pareto-based method, the Strength Pareto Evolutionary Algorithm (SPEA). The main features of this approach are: it (i) sorts the non-dominated solutions externally and continuously updates the population, (ii) evaluates an individual's fitness depending on the number of external non-dominated points that dominate it, (iii) preserves population diversity using the Pareto dominance relationship, and (iv) incorporates a clustering procedure in order to reduce the non-dominated set without destroying its characteristics.

Recently Pareto Differential Evolution (PDE) algorithm is developed by Sarker et al [21] to handle multiobjective optimization problems. In DE, each variable's value in the chromosome is represented by a real number. The approach works by creating a random initial population of potential solutions, where it is guaranteed, by some repair rules that the value of each variable is within its boundaries. An individual is then selected at random for replacement and three different individuals are selected as parents. One of these three individuals is selected as the main parent. With some probability, each variable in the main parent is changed but at least one variable should be changed. The change is undertaken by adding to the variable's value a ratio of the difference between the two values of this variable in the other two parents. In essence, the main parent's vector is perturbed by the other two parents' vectors. This process represents the crossover operator in DE. If the resultant vector is better than the one chosen for replacement, it replaces it; otherwise the chosen vector for replacement is retained in the population.

## III. PROPOSED APPROACH

EAs are stochastic search methods that have been applied successfully in many search, optimization, and machine learning problems. Unlike most other optimization techniques, EAs maintain a population of encoded tentative solutions that are manipulated competitively by applying some variation operators to find a satisfactory, if not globally optimum solution. Parallelism of evolutionary algorithm is performed to speed up the computation. NES algorithm [10] solve problem by using sequential approach. But it takes much time to achieve solution. In the proposed approach parallelization of the sequential NES algorithm is performed. The proposed distributed evolutionary approach is named as Distributed Novel Evolutionary Strategy Algorithm( DNESA). Fig 1. shows the approach of Multi-Objective optimization technique that is used in this paper. Multiple single objectives function are combined on the basis of high level information. Each single objective function are multiplied by a suitable weight factor. After that, functions are added and converted to single objective function. This single objective function is optimized by using the approach of single objective optimization.

### A. Distributed Novel Evolutionary Strategy Algorithm (DNESA)

The availability of powerful-networked computers presents a wealth of computing resources to solve problems with large computational effort. As the communication level in coarse-grained parallelization is low as compared to other parallelization strategies, it is a suitable computing model for distributed networks with limited communication speeds. A model of distributed computing is shown in Fig 2.

The parallelization approach is considered here, where large problems are decomposed into smaller subtasks that are mapped into the computers available in a distributed system.

The flowchart of the proposed DNESA is illustrated in Figure 3. The initialization process involves the creation of subpopulations of random individuals, where the $i^{th}$ subpopulation will represent the $i^{th}$ decision variable. The individuals will then undergo the competitive-cooperation process until the stopping criteria is satisfied, which can be set based upon a fixed number of function evaluations. In this work, the number of fitness function evaluations is determined according to past experience and complexity of the test functions, which can be in multiples of the number of decision variables.

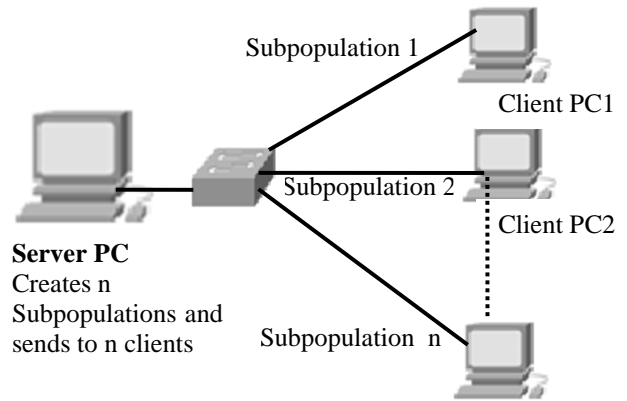

Figure 2. A model of DNESA.

Server generates the initial population and it also initializes the parameters such as generation number and size of the subpopulation. After that, the server calculates the initial fitness value of each individual so that individuals can be ordered and subpopulations can be generated. The condition is checked so that the generation number could not exceed maximum generation number. If so then the process is stopped and the result is reported. Subpopulations are generated according to the number of clients. Each subpopulation is assigned to each client and server waits for the result to come from the clients.

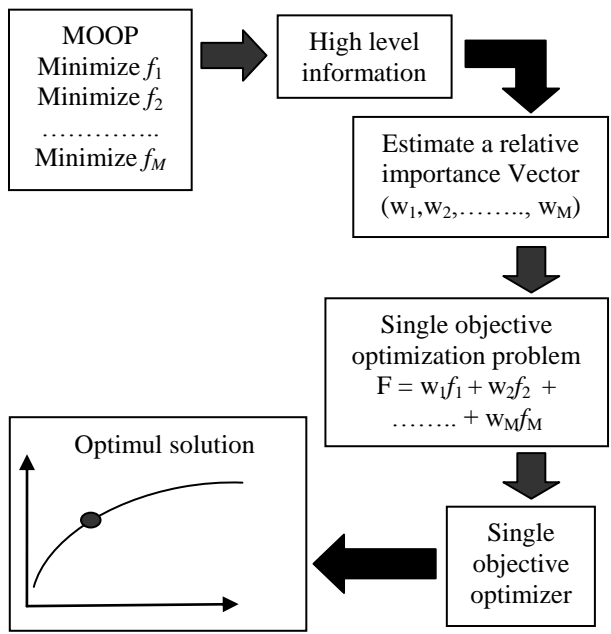

Figure 1. Approach of Multi-Objective Optimization

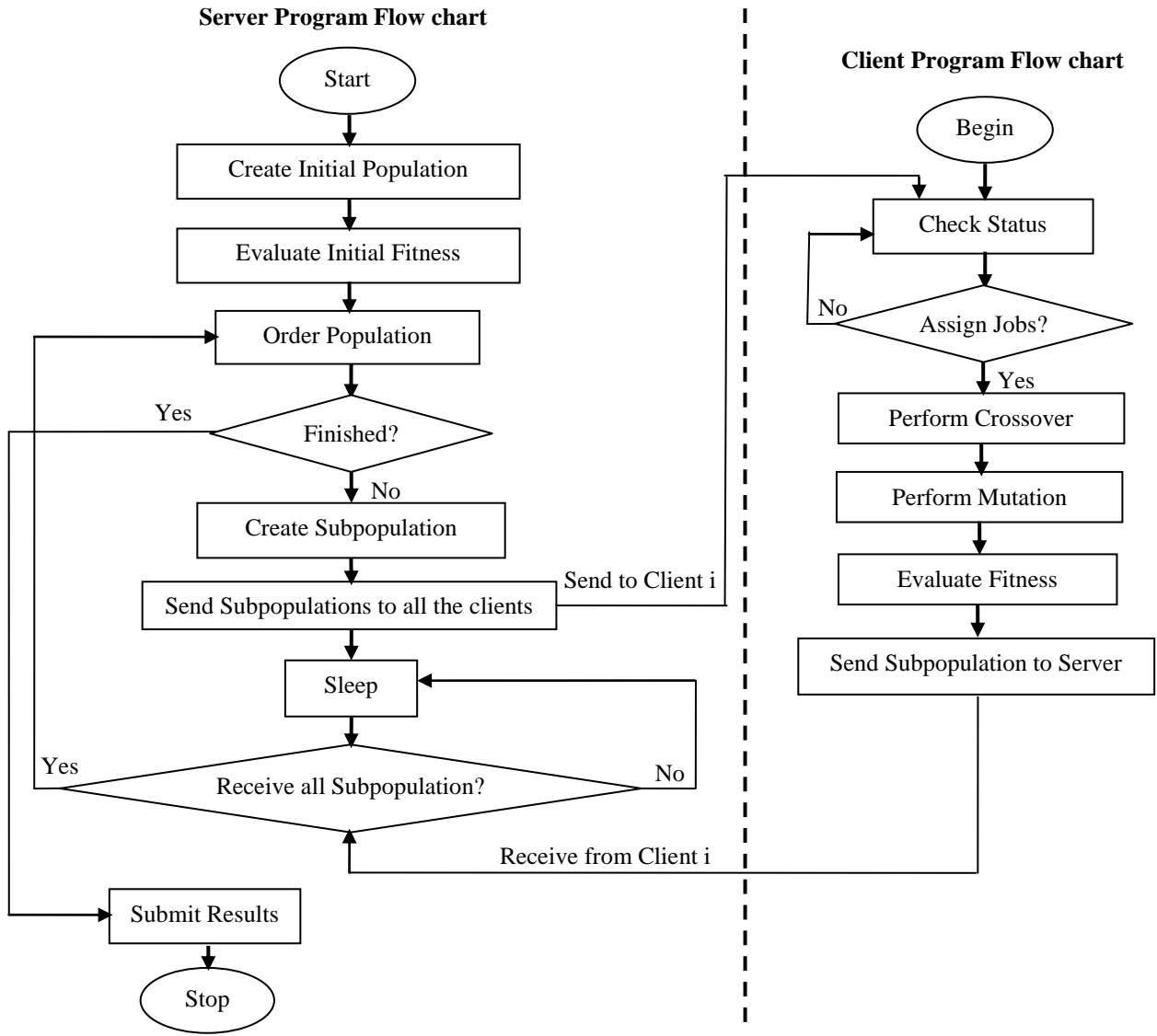

Figure 3. The program Flow chart of DNESA.

At the first step it finds out the representative and performs subpopulation based max min crossover (SBMAC) [10] operation to the representative for generating new subpopulation. After SBMAC, client performs time variant mutation(TMV) [10] for diversity among the generated individuals. Finesses are evaluated at the next stage. Then the fitness value and parameter values are returned to the server. The server checks that all subpopulations are received from clients, if so then it go back to order generation phase and repeat the operation until the generation number reach the maximum number of generation or the fitness value reached the expected goal.

In this parallel approach, for a particular generation all the subpopulations are send to the clients at a time and thus execution of each subpopulation is done at a time, which saves

## IV. EXPERIMENTAL SETUP AND RESULTS

Six benchmark problems are solved by using DNESA. The DNESA algorithm is implemented in the lab environment, where maximum 10 computers are used as clients and one computer as server. The communication between server and clients are established using Local Area Networking (LAN). Both the server and clients programs are written in C#.net framework and communication is done using socket programming. Initial population size, maximum number of generation varies problem to problem. The crossover rate and mutation rate are same for all problems. The initial population is generated by using a Uniform Random Number (URN) within desired domain of the object variables. After evaluating the fitness value of initial individuals, this population is considered as parents for the next generation. SBMAC [10] and TMV [10] are used for crossover and mutation respectively

TABLE I. DEFINITION OF TEST PROBLEMS

| Definition of Problems |
|---|
| Problem P1[10]:<br>$f_1(x) = x_1^2 + 2*x_2^2 - 0.3*\cos(3\pi x_1)\cos(4\pi x_2) + 0.3;$<br>$f_2(x) = x_1^2 + x_2^2 - 2.3*\cos(\frac{\pi x_1}{2})\cos(\frac{\pi x_2}{2}) + 0.3;$<br>where $-50 \le x_i \le 50$ for $i = 1,2$ |
| Problem P2[11]:<br>$f_1(x) = 1 - exp[-\sum_{i=1}^{n}(x_i - 1/\sqrt{n})^2];$<br>$f_2(x) = 1 - exp[-\sum_{i=1}^{n}(x_i + 1/\sqrt{n})^2];$<br>where $-2 \le x_i \le 2$ for $i = 1,2,3,\dots,n$. |
| Problem P3[12]:<br>$f_{11}(x) = x^2.$<br>$f_{12}(x) = (x-2)^2.$<br>where $-10 \le x \le 10$ |
| Problem P4[13]:<br>$f_1(x) = 1 - exp[-(x_1 - 1)^2 - (x_2 + 1)^2];$<br>$f_2(x) = 1 - exp[-(x_1 + 1)^2 - (x_2 - 1)^2];$<br>where $-10 \le x_i \le 10$ for $i = 1,2$ |
| Problem P5(DLTZ1)[9]:<br>$f_1(x) = \frac{1}{2}x_1 x_2 \dots x_{M-1}(1 + g(x_M));$<br>$f_2(x) = \frac{1}{2}x_1 x_2 \dots (1 - x_{M-1})(1 + g(x_M));$<br>$f_3(x) = \frac{1}{2}x_1 x_2 \dots (1 - x_{M-2})(1 + g(x_M));$<br>$\quad \vdots$<br>$f_{M-1}(x) = \frac{1}{2}x_1(1 - x_2)(1 + g(x_M));$<br>$f_M(x) = \frac{1}{2}(1 - x_1)(1 + g(x_M));$<br>$0 \le x_i \le 1$ for $i = 1,2,3,\dots,n;$<br>where $g(xM) = \sum_{x_i \in xM}(xi - 0.5)^2.$ |
| Problem P6(DLTZ2)[9]:<br>$f_1(x) = (1 + g(x_M))\cos(x_1\frac{\pi}{2})\cos(x_2\frac{\pi}{2})\dots\cos(x_{M-2}\frac{\pi}{2})\cos(x_{M-1}\frac{\pi}{2});$<br>$f_2(x) = (1 + g(x_M))\cos(x_1\frac{\pi}{2})\cos(x_2\frac{\pi}{2})\dots\cos(x_{M-2}\frac{\pi}{2})\sin(x_{M-1});$<br>$f_3(x) = (1 + g(x_M))\cos(x_1\frac{\pi}{2})\cos(x_2\frac{\pi}{2})\dots\sin(x_{M-2}\frac{\pi}{2});$<br>$\quad \vdots$<br>$f_{M-1}(x) = (1 + g(x_M))\cos(x_1\frac{\pi}{2})\sin(x_2\frac{\pi}{2});$<br>$f_M(x) = (1 + g(x_M))\sin(x_1\frac{\pi}{2});$<br>$0 \le x_i \le 1$ for $i = 1,2,3,\dots,n;$<br>where $g(xM) = \sum_{x_i \in xM}(xi - 0.5)^2.$ |

The proposed DNESA use server client model, where server controls multiple clients. It needs much less time to solve problems rather than single clients. When the number of clients increases, it reduce the time to find out optimal solution. Four problems (P1, P2, P3 and P4) in TABLE I are tested to find out the relationship between time needed to find optimal solution and number of clients. The experimented results are shown as plotted graph in Fig. 4 for the problems. From Fig. 4(a), it is shown that when single client is used for problem P1 then it takes much time to find optimal solution. But increasing the number of clients, it is shown that the time for finding the optimal solution is reduced. Also for problems P2 and P4 in Fig. 4(b) and Fig. 4(d), it is shown that increasing the number of clients reduced the execution time for finding optimal solution. But for problem P3 in Fig. 4(c) shows that initially time is reducing with increasing the number of clients, but after a certain number of clients it does not reduce time periodically. Also with the increasing clients, the ratio of the time reduction decreases due to extra time needed for the communication among the clients and server.

The DNESA is compared with some other MEAs to compare the performance of DNESA with respect to those algorithms. There are three other MEAs are used to compare with the proposed one. The MEAs are the PESA [6], NSGA-II [7] and SPEA2 [8]. Two benchmark problems are used for this comparison. The problems are DLTZ1 (P5) and DLTZ2 (P6). Each algorithm is compared with each other with respect to each problem. The algorithms are compared according to the convergence metric. Convergence represent the distance between the converged non-dominated solution and the global pareto-optimal front, hence lower values of convergence metric represent good convergence ability. Therefore, lower the value of convergence of an algorithm is more superior.

*A. Convergence for Problem DLTZ1 (P5)*

A lower convergence metric value implies better result. Following sections present results for convergence on the DLTZ1 test problem. The DNESA gives the better performance than PESA, SPEA2 and NSGA-II in terms of convergence for the DLTZ1 problem. PESA is better than SPEA2 and NSGA-II but not better than DNESA. SPEA2 and NSGA-II have similar performances. The data are taken for 20 and 50 individuals with a fixed generation of 300. From 30 runs with different sample path, the average value of convergence are calculated for the DNESA algorithm. The data for the PESA, SPEA2 and NSGA-II are recorded form data sheet of [9].

TABLE II. MEAN CONVERGENCE VALUE (CV) FOR PROBLEM DLTZ1

| Gen No. | Runs | Individuals | DNESA | PESA | SPEA2 | NSGA-II |
|---|---|---|---|---|---|---|
| 300 | 30 | 20 | 0.00129 | 2.86948 | 3.08825 | 2.27666 |
| | | 50 | 0.00128 | 0.04419 | 0.04843 | 0.38360 |

From the data TABLE II, it is shown that for 20 individuals convergence value of DNESA is much better than other algorithms. For 50 individuals also DNESA is the best among the algorithms.

*B. Convergence for Problem DLTZ2 (P6)*

DNESA gives the better performance than SPEA2 and NSGA-II but not better than PESA in terms of convergence for the DLTZ2 problem. PESA is better than NESA, SPEA2 and NSGA-II. SPEA2 and NSGA-II have similar performances. The data are taken for 20 and 50 individuals with a fixed generation of 300. From 30 runs with different sample path, the average value of convergence are calculated for the DNESA algorithm. The data for the PESA, SPEA2 and NSGA-II are recorded form data sheet of [9].

TABLE III. MEAN CONVERGENCE VALUE (CV) FOR PROBLEM DLTZ2

| Gen No. | Runs | Individuals | DNESA | PESA | SPEA2 | NSGA-II |
|---|---|---|---|---|---|---|
| 300 | 30 | 20 | 0.00138 | 8.40E-05 | 0.00026 | 0.00179 |
| | | 50 | 0.00083 | 0.00035 | 0.00663 | 0.01003 |

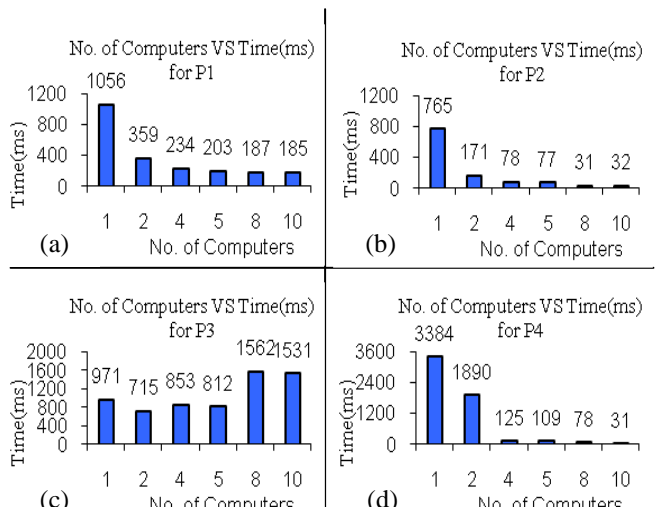

Figure 4.Time needed to find optimal solution with respect to number of clients for problems P1(a), P2(b), P3(c), P4(d).

From the data TABLE III, it is shown that for 20 individuals, convergence value of DNESA is much better than SPEA2 and NSGA-II but not better than PESA. For 50 individuals also DNESA is the better than SPEA2 and NSGA-II but little worse than PESA.

The DNESA is developed from NES algorithm and uses minimum genetic drift. Due to minimum genetic drift in DNESA, the DNESA shows better performance in convergence than other three algorithms.

## V. CONCLUSION

In this paper a distributed evolutionary computation technique for Multi-Objective Optimization problems is proposed. This distributed approach helps to achieve the optimal solution quickly for solving MOPs. In DNESA, total population is divided into subpopulations. This subpopulation concept makes easy to parallelize NES algorithm. Subpopulations are assigned to each client to find out optimal solution. All clients perform in parallel. Hence, the time to find the solution is greatly reduced. Four benchmark problems namely P1, P2, P3 and P4 are used to show the relationship between the numbers of clients versus the time needed to find out an optimal solution. DNESA is also compared with PESA, SPEA2 and NSGA-II for convergence metrics. Two benchmark problems namely P5 and P6 are used for comparison. The convergence value for the proposed DNESA is better than other three algorithms.

REFERENCES

[1] Carlos A. Coello Coello "A comprehensive survey of evolutionary-based multiobjective optimization techniques." Knowledge and Information Systems, VOL.1 NO. 3, pp. 269 - 308, 1998.

[2] Chankong V., Haimes, Y.Y "Multiobjective decision making theory and methodology". New Work: North Holland, 1983.

[3] Hans, A.E." Multicriteria optimization for highly accurate systems. Multicriteria Optimization in Engineering and Sciences", W. Stadler(ED.), Mathematical concept and methods in science and engineering, VOL. 19, pp. 309 - 352. New York:Plenum oress, 1988.

[4] Eckart Zitzler and Lothar Thiele. "Multiobjective Evolutionary Algorithms: A Comparative Case Study and the Strength Pareto Approach". IEEE Transactions on Evolutionary Computation, VOL. 3, NO. 4, pp. 257 - 271, November 1999

[5] Coello C A C." Evolutionary multiobjective optimization: a historical view of the field." IEEE Computational Intelligence Magazine ,VOL.1, NO. 1, pp. 28 - 36 ,February 2006.

[6] David W. Corne, Joshua D. Knowles, and Martin J. Oates. "The Pareto Envelope-based Selection Algorithm for Multiobjective Optimization". Proceedings of The Parallel Problem Solving from Nature VI Conference, pp. 839 - 848, Paris, France,Springer Publication, 2000.

[7] Kalyanmoy Deb, Samir Agrawal, Amrit Pratab, and T. Meyarivan." A Fast Elitist Non-Dominated Sorting Genetic Algorithm for Multi-Objective Optimization: NSGA-II." Proceedings of the Parallel Problem Solving from Nature VI Conference, pp. 849 - 858, Paris, France, Springer Publication, 2000.

[8] Eckart Zitzler, Marco Laumanns, and Lothar Thiele. " SPEA2: Improving the Strength Pareto Evolutionary Algorithm." May 2001.

[9] V.Khare, X. Yao and K.Deb, "Performance Scaling of Multiobjective Evolutionary Algorithms", In Proc, 2nd Int. Conf. Evolutionary Multi-criteria Optimization, pp. 376 - 390,Springer Publication, 2003.

[10] Hashem, M. M. A., "Global Optimization Through a New Class of Evolutionary Algorithm", Ph.D. dissertation, Diss. No. 19, Saga University, Japan, pp. 1 - 25, 1999.

[11] K. C. Tan, Y. J. Yang, and C. K. Goh, " A Distributed Cooperative Coevolutionary Algorithm for Multiobjective Optimization", IEEE Transactions on Evolutionary Computation, VOL. 10, NO. 5, pp. 527-549, October 2006.

[12] N. Srinivas and Kalyanmoy Deb, " Multiobjective Optimzation Using Nondominated Sorting in Genetic Algorithms" , Proceedings of Evolutionary Computation, Indian Institute of Technology, Kanpur, VOL. 2, pp. 221 248, 1994.

[13] Ehrgott M, "Multicriteria optimization." Springer, second edition, 2005.

[14] Schaffer, J, "Multiple objective optimization with vector evaluated genetic algorithms". Proceedings of the First International Conference on Genetic Algorithms, pp. 93 - 100, L. Erlbaum Associates Inc publisher,Hillsdale, NJ, USA, 1985.

[15] Hajela, P. and C. Lin , "Genetic search strategies in multicriterion optimal design". Structural Optimization, VOL. 4,NO. 2, pp. 99 - 107, June 1992.

[16] Goldberg, D.," Genetic algorithms: in search, optimisation and machine learning". Addison-Wesley Longman Publishing Co., Inc. Boston, MA, USA, 1989.

[17] Fonseca, C. and P. Fleming ," Genetic algorithms for multiobjective optimization: Formulation, discussion and generalization." Proceedings of the Fifth International Conference on Genetic Algorithms, San Mateo, California, pp. 416 - 423, 1993.

[18] Horn, J., N. Nafpliotis, and D. Goldberg ," A niched pareto genetic algorithm for multiobjective optimization." Proceedings of the First IEEE Conference on Evolutionary Computation VOL. 1, pp. 82 - 87, 1994.

[19] Knowles, J. and D. Corne ," The pareto archived evolution strategy: a new baseline algorithm for multiobjective optimization." In 1999 Congress on Evolutionary Computation, Washington D.C., IEEE Service Centre, pp. 98 - 105, 1999.

[20] Knowles, J. and D. Corne ," Approximating the nondominated front using the pareto archived evolution strategy." Evolutionary Computation, VOL. 8, NO. 2, pp. 149 - 172. MIT Press Cambridge, MA, USA, 2000.

[21] Abbass, H., R. Sarker, and C. Newton ." A pareto differential evolution approach to vector optimisation problems." Congress on Evolutionary Computation 2, 971–978. 2001